# Force Field Generalization and the Internal Representation of Motor Learning

Alireza Rezazadeh* and Max Berniker

Department of Mechanical and Industrial Engineering,
University of Illinois at Chicago, Chicago, IL, USA

*Corresponding Author: Alireza Rezazadeh, arezaz2@uic.edu

*Abstract*— When learning a new motor behavior, e.g. reaching in a force field, the nervous system builds an internal representation. Examining how subsequent reaches in unpracticed directions generalize reveals this representation. Though it is the subject of frequent studies, it is not known how this representation changes across training directions, or how changes in reach direction and the corresponding changes in limb impedance, influence measurements of it. We ran a force field adaptation experiment using eight groups of subjects each trained on one of eight standard directions and then tested for generalization in the remaining seven directions. Generalization in all directions was local and asymmetric, providing limited and unequal transfer to the left and right side of the trained target. These asymmetries were not consistent in either magnitude or direction even after correcting for changes in limb impedance, at odds with previous explanations. Relying on a standard model for generalization the inferred representations inconsistently shifted to one side or the other of their respective training direction. A second model that accounted for limb impedance and variations in baseline trajectories explained more data and the inferred representations were centered on their respective training directions. Our results highlight the influence of limb mechanics and impedance on psychophysical measurements and their interpretations for motor learning.

## I. Introduction

Learning a new motor behavior entails building an internal model [1-3]. This internal model, or representation, allows us to perform the learned task now and at future times. Importantly, it also enables us to generalize what we have learned to new, unpracticed circumstances. Experimental studies exploit this ability to measure the internal representation of a motor task. For example, after learning to make a reach in a force field, subjects are asked to reach to new directions [4-7]. The resulting movements and forces subjects apply in these new directions are used to infer their representation of the learned force field. This standard approach is frequently used to examine the internal representation in a variety of motor tasks including movements with force [8-10] or visual perturbations [11, 12]. Prior work on velocity-dependent force fields has found narrow generalization curves, indicating adaptation in a single direction is local and only benefits movements made in nearby directions, usually no more than 45° from the trained direction [5, 6, 8, 13-15]. Generalization curves are often asymmetric as well, indicating that the representation of a learned force field does not decay symmetrically to the left or right side of the trained direction [7, 10]. Differing explanations have been offered in terms of whether the learned representation is associated with the intended, or actual, reaches made [8], or whether variations in motor errors due to limb stiffness can account for these changes [7]. Still other studies do not find asymmetries [6]. As such, many important questions regarding the learned representation remain unanswered.

Despite the many studies examining generalization, previous work has mostly been limited to a small number of reaching conditions; usually training in a single standard direction. Whether or not the aforementioned findings hold for other reach directions remains unverified. Additionally, how the mechanics of the limb itself influence these generalization measurements has not been examined. Changes in limb impedance, either due to changes in movement direction or due to newly adapted motor commands, will necessarily influence the dynamics of the limb [16, 17]. It is not clear how these changes may influence generalization measurements and in turn our estimates of the internal representation.

We performed a force field adaptation experiment using eight groups of subjects each practicing movement in one of eight directions. After adapting we measured their ability to generalize in the remaining seven directions using error-clamps. Examining adaptation and generalization in all eight directions allowed us to correct for changes in limb impedance across reach directions. We found local and asymmetric generalization curves in all eight directions. However, these asymmetries were not consistent in magnitude, strength, or direction, calling into question previous explanations for how the internal representation is acquired. Using the standard assumption that adapted forces are a "read-out" of the representation (the estimated force field strength), we found that they inconsistently shifted to the left or right side of the training directions. A second model accounting for limb dynamics and the details of unperturbed trajectories was able to explain both baseline and post adaptation forces. The inferred representations were more consistent and centered on the training directions. Our results highlight the influence of limb mechanics on psychophysical measurements, and how these influences can alter interpretations of motor learning.

## II. Methods

### A. Subjects

80 right-handed subjects (25±4.4 years old, 21 females) participated in this experiment. Right handedness was assessed using the Edinburgh inventory [18]. All subjects had normal or corrected-to-normal vision, no history of motor deficits and were naïve to the purpose of the study. All experimental protocols were approved by University of Illinois at Chicago Institutional Review Board. Subjects provided written informed consent before participation and were compensated using a baseline salary and bonus based on performance (for a minimum of $10 and a maximum of $20).

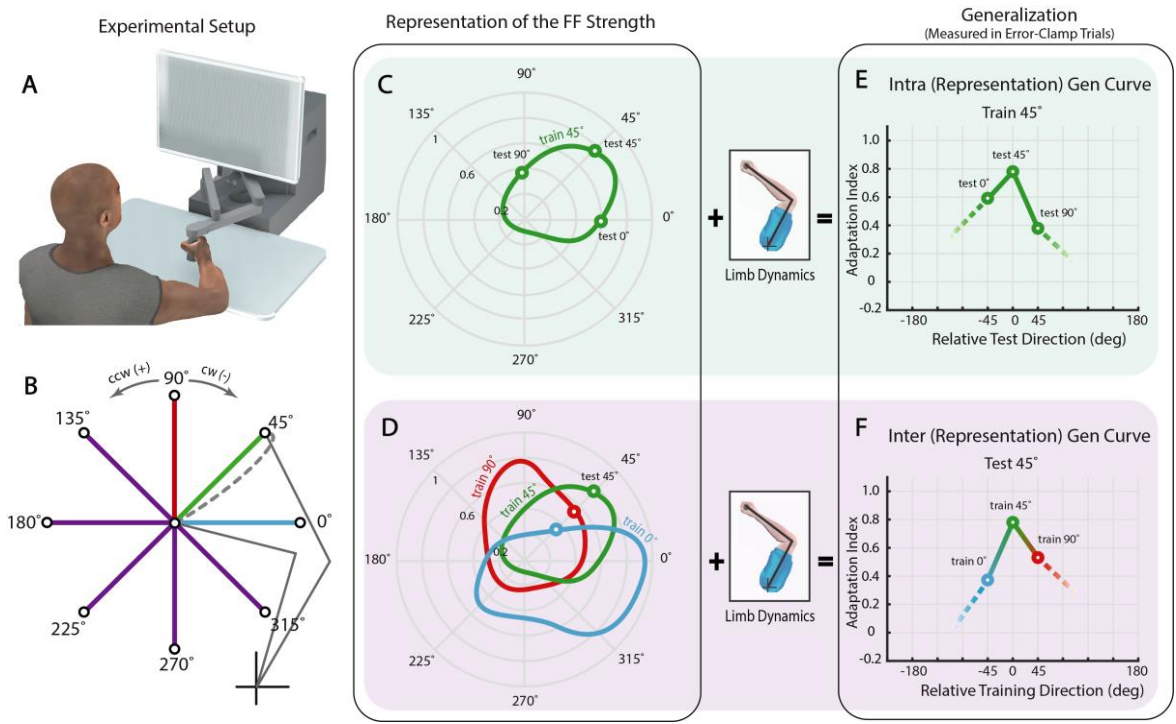

**Fig 1. Experimental Setup and Generalization Curves.** (A) Subjects made reaches in a null field, a curl field or an error-clamp. After landing on target the robot brought their hand back to the center position. (B) Eight standard targets were located 10-cm away from a home target. Each of eight groups practiced reaching in a CW force field towards one of the eight targets (the gray dashed line depicts a reach in the field to the 45° target) and then tested for generalization to all remaining targets using error-clamp trials. (C) A schematic of the internal representation of the learned force field for the 45° target group. The curve depicts a subject's estimated strength of the force field in polar coordinates (green contour). (D) Using this estimate the limb makes a generalization reach. Combining the measurements made from many such reaches after training on the 45° target constitutes an intra-representation generalization curve (green circles). These measurements are influenced by both the force field estimate and changes in limb impedance with reach direction. (E) Schematic of representations from three groups of subjects learning the field for the 0° (blue contour), 45° (green contour), and 90° (red contour) targets. (F) Combining the measurements made across all three groups when reaching to the same 45° target constitutes an intra-representation generalization curve. This curve corrects for the changes in reach direction.

*B. Experiment Overview*

The Subjects held the handle of a robotic manipulandum with their right hand while seated in a height-adjustable chair in front of a vertical LCD display (Fig 1A). The location of the chair was adjusted for each subject such that when their limb was in the center of the workspace it was in a standard configuration (shoulder angle of 45° and a relative elbow angle of 90°). The chair height was adjusted to keep the upper and lower arm approximately in a plane aligned with the robot handle. Their upper and lower arms were suspended from the ceiling using height-adjustable sling to limit the effect of gravity and prevent fatigue.

All trials began with the subject's hand resting in a center position (the home target), until a target appeared on the screen. Subjects were instructed to reach with one fast, smooth motion to the target as soon as it appeared. Once inside the target and after a small delay the robot smoothly returned their hand back to the center position. After a small random delay, the next trial would begin.

All Targets were 10-mm diameter circles, spaced 45° apart, on a 10-cm radius circle centered on the home position (Fig 1B). Visual feedback of the hand was provided with a 10-mm diameter white cursor accurately depicting the hand's relative displacement in the workspace. Successful trials landed within the target and within a specified time window (300-450ms). To provide feedback on performance, visual and audio feedback were given after each reach. Targets were initially displayed as yellow circles, but changed to blue, red, or green when the reach was too slow, too fast, or within the desired time, respectively. Once they came to a stop (speed<2cm/s) an auditory tone was emitted: 1KHz if too slow, 20KHz if too fast and 10KHz if on time. Successful reaches were awarded a point and their running score was advanced on the screen's display.

On each trial the robot would provide either a force field or catch trials with either a null field or an error-clamp. The force field was a velocity-dependent clockwise curl field (15Ns/m). The null field attempted to rendered zero forces at the handle of the robot. Error-clamp trials constrained the hand's motion along a straight channel connecting the home and target locations. In addition to these three force conditions, each trial either provided continuous visual feedback of the hand location, or the hand's location was extinguished when the target appeared and didn't appear again until the hand came to a complete stop.

## C. Experimental Protocol

Eighty subjects were randomly assigned to eight groups, each practicing reaches in the force field to one of eight training directions (0°, 45°, 90°, 135°, 180°, 225°, 270°, 315°). The experiment was divided into four blocks. The 1st block (baseline) was to familiarize subjects with the task and collect baseline measurements. 208 trials (26/target) to all eight targets were presented in a pseudo-random order. Trials were made under the null field condition with feedback of the hand location, except for a randomly ordered 24 trials (3/target) that were error-clamp trials without visual feedback of the hand.

Next were two identical adaptation blocks (2nd and 3rd), where subjects practiced reaching to their assigned training target repeatedly. These were either force field trials or catch trials with an error-clamp to assess adaptation and performance. Each block was 65 trials, 60 of which were force field trials, of which 15 had the feedback extinguished. The remaining 5 trials were error-clamps without visual feedback. By extinguishing feedback randomly during force field and catch trials we eliminated the chance of subjects predicting the error-clamp trials. Subjects had a short rest period (3-5mins) between the two blocks of adaptation.

The 4th block (testing) was used to measure the ability to generalize the learned force field to new reach directions. 210 reaches were made, repeatedly switching from the training target to one of the seven remaining test targets. This allowed us to test a subject's ability to generalize the learned force field to all directions while preventing washout of the adapted behavior. All 105 reaches (15/target) to test targets were made in error-clamps without visual feedback. Of the remaining 105 trials with the train target, 53 had the field and feedback of the hand, 26 had the field and no feedback, and 26 were error-clamp trials without feedback. As with blocks 2 and 3, by mixing force and feedback conditions we ensured subjects could not distinguish between error clamp and force field trials.

## D. Experimental Apparatus

Reaches were made with a two-joint robotic manipulandum (KINARM End-point robot, BKIN Technologies, ON, CA) with custom software written explicitly for this experiment running in real-time at 2KHz. Hand location and end-point forces (using a transducer at the handle) were measured and stored for post-processing at 1KHz.

The curl field was defined as,

$$\begin{bmatrix} F_x \\ F_y \end{bmatrix} = \alpha \begin{bmatrix} 0 & 1 \\ -1 & 0 \end{bmatrix} \begin{bmatrix} v_x \\ v_y \end{bmatrix} \quad (1)$$

where Fx and Fy are the x and y-components of the field and vx and vy are the hand's velocity, and α=15Ns/m, is the strength of the clockwise force field. The error-clamp was rendered by creating a virtual channel along a straight line between the home position and the target [19]. The width of this channel was 1mm and bounded on either side by stiff walls created with a virtual spring (5KN/m) and damper (5Ns/m), that only pushed the hand inwards towards the channel.

To render an accurate desired force at the handle of the robot, a low-gain force-feedback loop was used to compensate for the manipulandum's inertia. The commanded forces were

$$F_{robot} = F_{desired} + K(F_{desired} - F_{measured}) \quad (2)$$

where K was 0.5 for the null field and 0.75 for the force field.

## E. Data Analysis

The velocity of the hand was computed by discretely differentiating the measured hand position and then smoothing with a 50Hz, 2nd order low-pass Butterworth filter. Using the hand's velocity, a starting (velocity> 5cm/s) and stopping time (velocity < 2cm/s) were defined for each trial. Also, for each trial the maximum perpendicular error (PE) was computed as the maximum deviation of the hand's path from a straight line connecting the reach starting location and the target location.

Adaptation indices were computed to measure how accurately subjects predicted the force field. For each error-clamp trial, forces applied to the walls of the channel were linearly regressed against the correct forces associated with the hand's velocity. This regression was performed on data within a time window defined by the trial's start and stop times (see above). This restricted the regression to the ballistic portion of the movement ignoring any deliberate or corrective movements that may have occurred during the later portion of the reach. The result of this regression was then divided by the field's actual strength, 15Ns/m. The index is equal to one if the channel forces exactly matched the force field profile, and zero if there were no forces applied to the channel walls.

Using these indices, we computed two generalization curves. The traditional curve depicts how the indices vary as subjects reach to test targets rotated away from the training direction. We refer to these curves as intra-generalization (intra-gen) curves (Fig 1E). To control for changes specific to reach directions, we also present curves that depict how adaptation indices vary as the training direction rotates away from a fixed test target. We refer to these as inter-generalization (inter-gen) curves (Fig 1F).

## F. Modeling the Learned Representation

Using As is common, the dynamics of the limb were modeled as a rigid two-link mechanism (Fig 1B), accelerated by self-generated commanded torques (τ), and the experimental forces (Fexp),

$$I(q)\ddot{q} + C(q,\dot{q})\dot{q} = \tau + J^T F_{exp} \quad (3)$$

where q is a vector of shoulder and elbow angles, I(q) is the inertia matrix, C(q, q̇) is a matrix of Coriolis and centripetal accelerations, and J is the end-point Jacobian. The arm's parameters were selected based on averages reported in previous work (m1 = 1.93 Kg, m2 = 1.52 Kg, r1 = 0.165 m, r2 = 0.19 m, l1 = 0.33 m, l2 = 0.34 m) [3, 20].

*a) The Standard Model:* A common assumption in force field generalization studies is that subjects intend to move along straight paths and the forces measured in error-clamp trials are a direct read-out of what subjects have learned; that is, forces are the subject's estimate of the force field [9, 10, 21]. The same reasoning suggests that channel forces measured in baseline should be close to zero, since the subjects are not yet aware of a field, and any deviations from it are due to noise or some constant offset. Therefore, these baseline measurements (pre-adaptation) are subtracted from those made in error-clamp trials after training (post-adaptation), to measure a subject's learned representation of the force field.

This representation is often quantified in terms of the measured adaptation index. Since the adaptation index decays as the reaching direction changes relative to the trained direction, the conventional form for modeling this representation is a Gaussian-like function:

$$\hat{\alpha}_j(\theta) = A_j \exp\frac{-(\theta - \theta_j - \mu_j)^2}{2\sigma_j^2} \quad (4)$$

where θ is the reach direction, and the free parameters of this model are the amplitude (A), the width (σ), and an offset (μ) relative to the training direction (θj). The index, j, varies from one to eight, to account for the eight training groups and their respective targets. Thus, the internal representation is assumed to be an estimate of the force field's strength, α̂, which varies over changes in reach direction relative to the training direction.

We assume the self-generated torque (τ) in Eq1 is a feed-forward command that drives the limb along a nominal minimum jerk trajectory while compensating for the estimated field.

$$\tau = I(q)\ddot{q}_{desired} + C(q,\dot{q})\dot{q}_{desired} + J^T \hat{\alpha} \begin{bmatrix} 0 & 1 \\ -1 & 0 \end{bmatrix} \begin{bmatrix} v_x \\ v_y \end{bmatrix} \quad (5)$$

Note that since the desired trajectory is always a straight line, feedback terms would not influence the movement or forces in the error clamps, and are thus not included. Furthermore, according to this model and its assumption of desired straight reaches, the baseline corrected channel forces should be a direct read-out of the estimated force field.

*b) The Limb Impendace Model:* Channel forces measured in the error-clamp trials, while likely containing an estimate of the force field, may also be influenced by the mechanics of the limb itself. If a subject's intended movement is not perfectly straight channel forces are a combination of the forces produced to counter the field and the limb's impedance. Recognizing this, we sought an alternative model to infer the learned representation.

This In this second model we assumed the self-generated torque is composed of two components, a feed-forward term to compensate for the limb's own dynamics and the estimated field, and a feed-back term to keep the limb stable when perturbed, τ = τff + τfb where,

$$\tau_{ff} = I(q)\ddot{q}_{desired} + C(q,\dot{q})\dot{q}_{desired} + J^T \hat{\alpha} \begin{bmatrix} 0 & 1 \\ -1 & 0 \end{bmatrix} \begin{bmatrix} v_x \\ v_y \end{bmatrix} \quad (6)$$

$$\tau_{fb} = K(q - q_{desired}) + B(\dot{q} - \dot{q}_{desired}) \quad (7)$$

The nominal values for the feedback gains were determined from impedance of the arm reported in previous work [3, 20], but scaled by constant terms to account for changes before and after adaptation; stiffness: K = αk [32, 16; 16, 21]Nm/rad, damping: B = αb [5,3; 3,4]Nms/rad.

Importantly, we made two additional assumptions. First, we assumed the representation for the learned field had no offset, but was instead centered on the learning direction,

$$\hat{\alpha}_j(\theta) = A_j \exp\frac{-(\theta - \theta_j)^2}{2\sigma_j^2} \quad (8)$$

Second, we did not assume that subjects' nominal reaches were perfectly straight. Instead, we assumed the average baseline movements were the desired movements, qdesired (t) = qbaseline (t). For this model the free parameters are the amplitude (A), and width (σ) of the representation, and two scaling factors for the stiffness and damping matrices (αk and αb).

With each model, we could simulate reaches and their channel forces in either baseline or generalization trials. The resulting forces could then be regressed to obtain simulated adaptation indices for each training group. Parameters for both models were found by minimizing the negative log-likelihood of the adaptation index data. Model comparisons were made using the corrected Akaike Information Criterion (AICc) and the Root-Mean-Square error (RMSE). Significance levels were set to 0.05 for all statistical tests. All code was custom written in MATLAB.

### III. RESULTS

To probe the underlying representation of a learned force field, subjects were randomly assigned to learn a curl field in one of eight standard directions and then tested for the ability to generalize in the remaining seven directions using error-clamp trials. The results were used to build standard generalization curves and new generalization curves that control for changes in reach direction. These curves were then used to infer the underlying representation of a newly learned force field.

All subjects began by performing a baseline block to characterize their unperturbed trajectories and initial error-clamp forces. Reaches were made in a null field with and without visual feedback, or error-clamp trials without feedback (see Methods). Not surprisingly, reaches made with visual feedback were relatively straight (Fig 2A). Perpendicular errors, though relatively small (ranging from 3.66 to 11.58mm) were not uniform across reach directions (one-way ANOVA, $F_{(7,10923)} = 169.7$, $p < 10E-16$, see Fig 2B). Errors were larger in directions of small arm stiffness (45° and 225°) and smaller in directions of large stiffness (135° and 315°). This pattern of errors is consistent with the long-held notion that the limb's stiffness can assist in producing straight paths [22, 23].

Baseline error-clamp trials, measuring endpoint forces before subjects were exposed to the force field served as a point of reference from which to gauge subsequent adaptation. Averaging across the eight groups of subjects we found baseline indices that varied considerably and were significantly different from zero for several directions (e.g. targets 45° and 225°, Fig 2D). Across reach directions, baseline perpendicular errors and adaptation indices were strongly correlated ($r = 0.93$, $p = 7E-4$). This finding is likely a direct result of the reach kinematics, since movements with nominal deviations from a straight line should elicit non-zero adaptation indices.

Perhaps more interesting, however, was the finding that all the adaptation indices that were significantly different from zero were also positive. This was generally true for all groups and most subjects (see supplemental Fig S1), despite the fact that subjects were making unperturbed reaches and had not yet been exposed to the field.

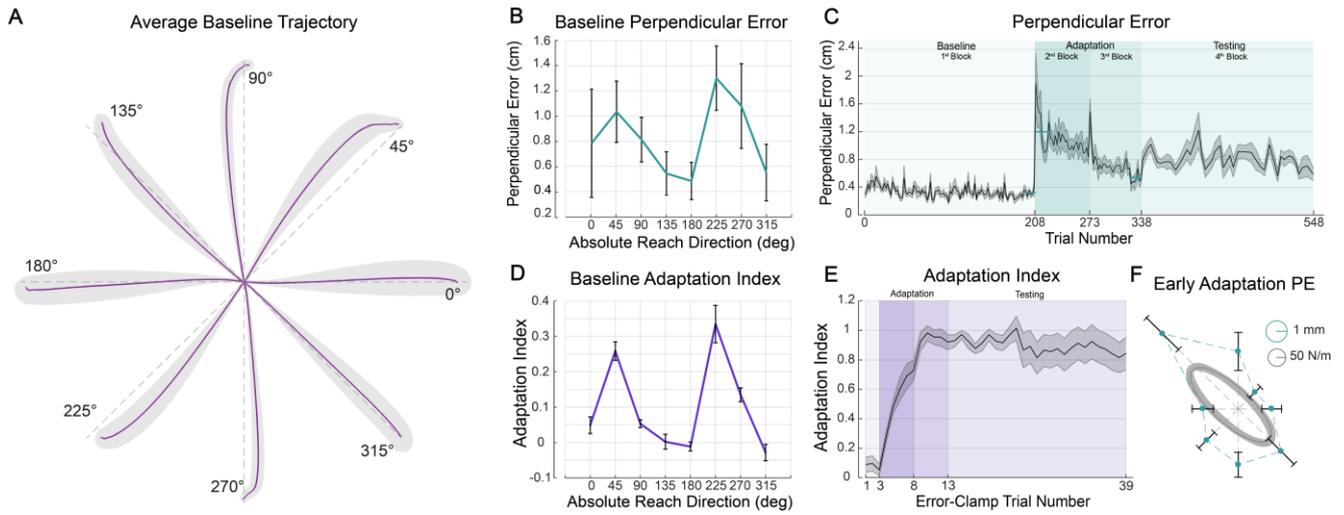

**Fig 2. Experiment's behavioral results.** (A) Baseline reaches in null field made with visual feedback (across-subject averages ± SEM). Reaches slightly deviate from a straight path (gray dashed lines). (B) Baseline perpendicular errors averaged across subjects. (C) Across-subject average perpendicular errors (±SEM) are plotted across all trials of the experiment. Only error-clamp trials are excluded. Subjects practiced reaching in force field during the 2nd and 3rd blocks with a brief rest between. (D) Baseline adaptation index averaged across subjects. Note the qualitatively similar shape to the perpendicular errors in B. (E) Force field learning curve averaged across all subjects and all reach directions. Mean values to the training targets (±SEM) are plotted across error-clamp trials. (F) Early adaptation perpendicular errors (mean ±SEM of first 20 trials) are averaged across subjects and overlaid on a hand stiffness ellipse from [7]. Note the qualitatively similar shapes of errors and hand stiffness.

After baseline, subjects practiced reaching in the curl field with two blocks of 65 trials (130 trials in total). As expected, early reaches were curved with typical "hook" shaped trajectories, but by the end of the second adaptation block were relatively straight. To quantify these changes in trajectories we compared perpendicular errors. Early errors were significantly larger than baseline (averaged last 20 trials in baseline compared with the averaged first 20 trials in the force field across all subjects: paired t-test, t (79) = 16.32, p < 10E-26). The errors gradually decreased although no clear plateau in this metric was observed (Fig 2C). Late training perpendicular errors were not similar to those made in the late baseline (averaged last 20 trials in the field compared with the averaged last 20 trials in baseline: t (79) = 6.36, p = 1.19E-5). Adaptation indices gradually increased to an average of 0.92 ± 0.05 (Fig 2E). Therefore, despite no clear plateau in perpendicular errors, force profile's indicated a near complete compensation for the field, somewhat higher than similar adaptation studies [24, 25].

Comparing errors across groups, there was an evident directional difference in the movement curvature during early adaptation (Fig 2F). As has been observed elsewhere, movement errors were lower when the arm was perturbed in the directions of larger limb stiffness [7]. Along with the baseline adaptation indices, this is further evidence of the role limb impedance plays in adapted behavior and the importance of taking this into account when interpreting learning.

Next, subjects performed a test block measuring their ability to generalize. Subjects repeatedly reached to their training target and then one of the seven remaining targets in error-clamp trials. Repeated switching from the test target to the training target allowed subjects to retain their familiarity with the force field throughout this block. Although a qualitative change in the adaptation indices SEM can be observed halfway through this block, no significant change between the first and last indices could be found (paired t-test, t(79) = 0.21, p = 0.83, Fig 2E)

Error-clamp data from these two test blocks was used to compute generalization curves, which ultimately were used to infer subjects' learned representations of the field. To this end, two generalization curves were created. Intra-representation generalization (or intra-generalization for brevity) curves, quantify the extent to which learning generalizes from a trained direction, to new (test target) directions (Fig 3A, see also Fig S2). Though these are the conventional curves for measuring generalization, combining measurements across changes in reach direction necessarily requires changes in limb posture and impedance. It is not clear if or how these changes in limb impedance influence the resulting force channel measurements. With this in mind a second set of generalization measurements were made, mitigating the influence of changes in limb impedance by only comparing reaches to the same target. Inter-representation generalization (inter-generalization) curves quantify the extent to which learning generalizes to a single test direction, after training in one of any neighboring (train target) directions (Fig 3B, see also Fig S3). With these two sets of curves we were able to perform a thorough examination of generalization and the possible effects due to limb impedance.

Intra-generalization curves, less their baseline contributions, were local; adaptation to the force field provided little assistance to movements separated by more than 45° from a training direction (Fig 3A). As has been reported elsewhere [7, 10], these curves were also asymmetric; the ability to generalize a learned force field did not decay equally in the left and right side of the trained direction. We quantified this asymmetry by computing the difference in the neighboring adaptation indices (-45° from +45°) for each training direction (Fig 3A).

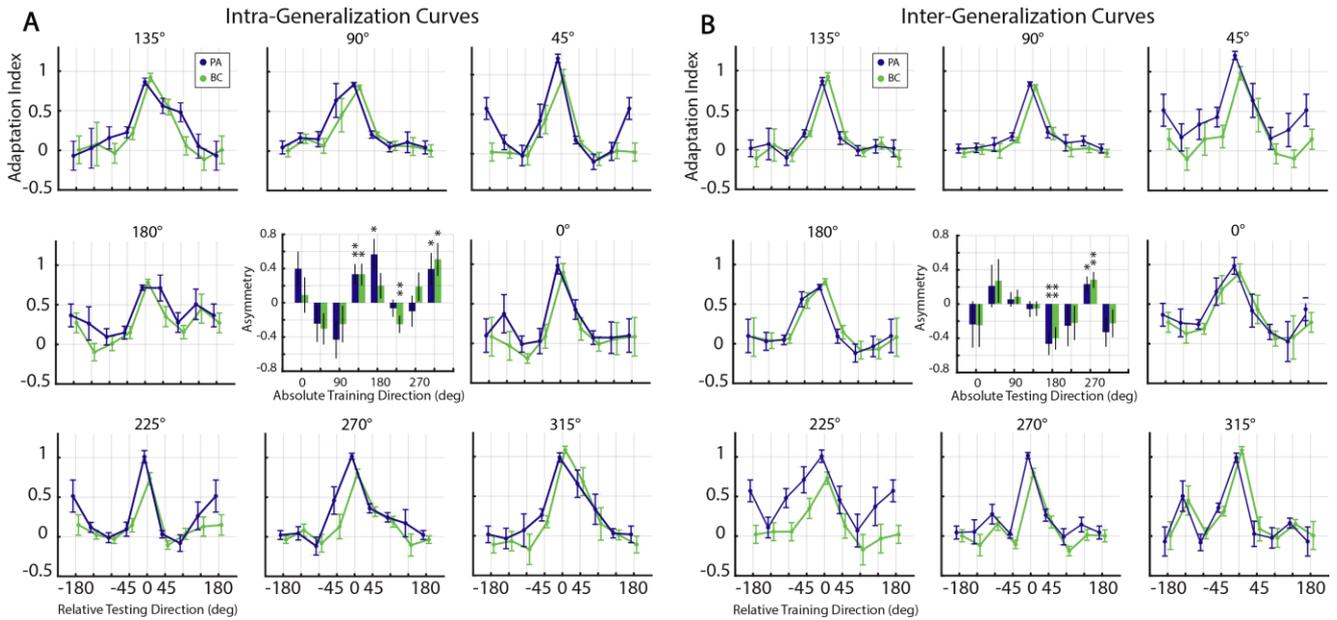

**Fig 3. Generalization of the velocity dependent force field.** (A) Intra-generalization curves (across-subject average ±SEM), quantifying the extent that learning the force field in a single training direction generalizes to all neighboring test directions. Each panel displays post-adaptation (PA) indices measured in the test block (green curves) along with baseline corrected (BC) adaptation indices (blue curves) as a function of relative angular distance of test targets relative to the training target. Adaptation indices for ±180° are identical. Asymmetry for each training direction is quantified as the difference between the neighboring adaptation indices (-45° from +45°) for post-adaptation and baseline corrected data (center panel). (B) Inter-generalization curves, quantifying the extent that learning the force field in any single neighboring direction generalizes to a testing direction. Each panel is displays post adaptation (PA) indices measured in the test block (green curves) together with baseline corrected (BC) adaptation indices (blue curves) as a function of the relative angular distance from the test target. Asymmetry for each curve is also shown (center panel). Asterisks indicate significant differences from zero (* = p < 0.5, ** = p < 0.01).

The asymmetries for each training direction were different in size and magnitude. That is, although learning the force field in some directions (such as upright, 90°) generalized more to the counter clock-wise neighboring target, learning in other directions (such as the upper left, 135° target) generalized more to the clock-wise adjacent target. The asymmetry was significant for three of the eight training directions in the post-adaptation intra-gen curves (two sample t-test, 135°: t(18)=2.8850, p=0.0099, 180°: t(18)= -2.5476, p=0.0202, 315°: t(18)= 2.1100, p=0.0174). Baseline corrected intra-gen curves were also significantly asymmetric for three out of eight training directions, two of which were similar to the uncorrected curve's directions (two-sample t-test, 135°: t(18)= 2.6386, p= 0.0167 ,225°: t(18)= 2.6181, p=0.0057, 315°: t(18)= 2.1100, p=0.0491). Although we observed an asymmetry similar to that previously reported for the 0° training target [8, 10], in general these asymmetries were not sufficiently consistent for an obvious interpretation.

After Inter-generalization curves were also local and asymmetric (Fig 3B). As with the intra-gen curves, the distance over which adaptation benefits movements to new directions fell off sharply after 45°. Inter-generalization asymmetries indicate that adapting to one side of a reach may be more beneficial than adapting to the other side. The asymmetries were inconsistent in size and magnitude (Fig 3B). Post-adaptation inter-gen curves were significantly asymmetric for only two out of eight test directions (two-sample t-test, 180°: t(18)=3.6035, p=0.0020, 270°: t(18)=-2.5960, p=0.0183). Also, baseline corrected intra-gen curves were significantly asymmetric in the same directions (two-sample t-test, 180°: t(18)= 3.0611, p=0.0067 , 270°: t(18)= -3.0808, p=0.0064). As with the previous intra-gen curves, there was no clear indication as to explain the inter-generalization asymmetries. Despite the differences in the asymmetries, both the intra- and inter-generalization curves were largely similar, suggesting that the uncontrolled changes in limb impedance did not have a qualitative influence on their measurements.

We then used the generalization data to infer the underlying representations subjects obtained through adaptation. First, we used a standard interpretation for the data. Assuming subjects intend to make perfectly straight reaches, and compensate precisely according to their estimate, the channel forces are a direct read-out of a subject's estimate of the force field (though equal and opposite). A model coupling the limb and the subject's representation could simulate error-clamp trials, the resulting forces and predicted adaptation indices (see Standard Model, Methods). Note that according to this standard interpretation the model's prediction for baseline adaptation indices are identically zero. However, by fitting the post adaptation predictions to average subject data, we could then infer each group's representation of the field. This was done for each of the eight groups by maximizing the log-likelihood of the adaptation index. The resulting representations were the subjects' average estimates of the force field strength as a function of reach direction.

This first model accurately fit the baseline corrected generalization data, with low error and a high goodness of fit (see Fig 4B, RMSE=0.107, AICc=-73.08). Naively we anticipated these representations would be largely the same for each group, since all the subjects adapted to the same force field, and the generalization curves were grossly similar.

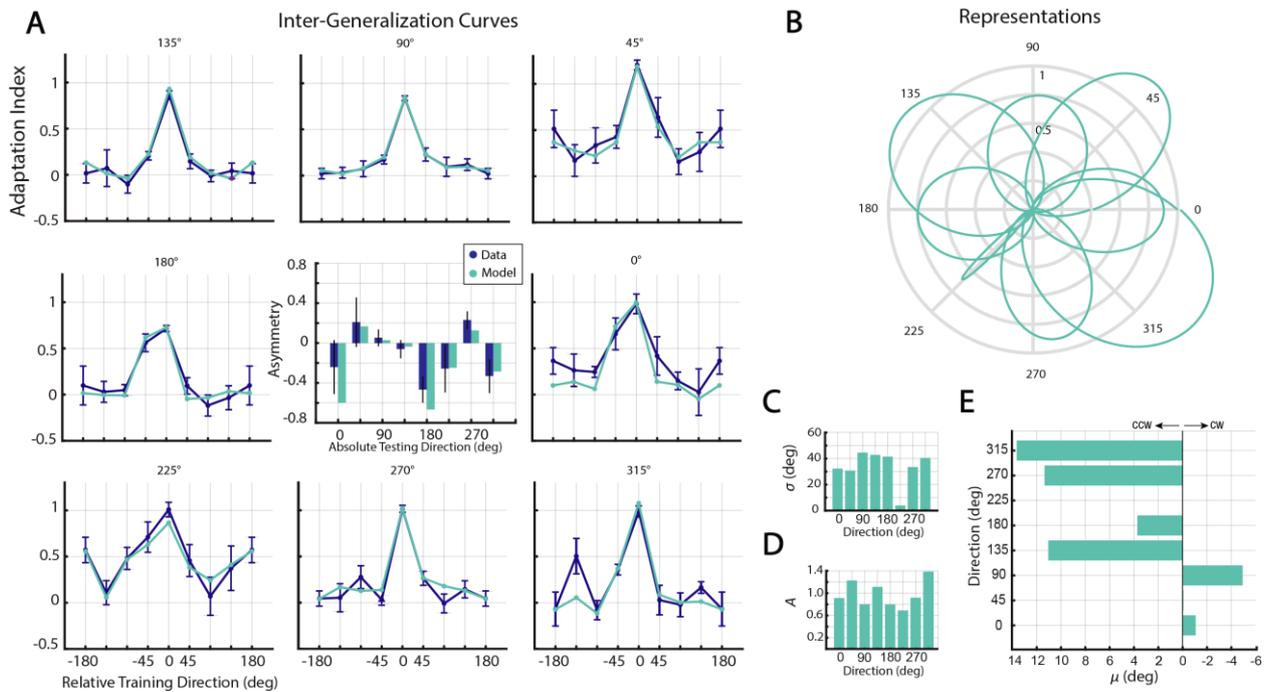

**Fig 4. Standard model results.** (A) Experimental and model-predictions for post-adaptation inter- generalization data (blue and cyan respectively). Experimental baseline indices are added to the model's predictions for comparison. Predicted asymmetries alongside their experimental counterparts (center panel). (B) The inferred representations are shown in polar coordinates for each of eight directions. The representation widths, σ (C) magnitudes, A (D) and offsets, μ (E) for each training direction are shown. Note the inconsistent shift of the representations to the clockwise and counterclockwise side relative to training directions.

However, by fitting small variations in the generalization curves the inferred representations had many differences (Fig 4C-E).

The "width" of the representations varied from group to group (average 33.67°±13.10°), the widest (90° target) being 44.53°, and the narrowest (225° target) being only 3.93°. The center of these representations was similarly variable (average 4.22°±6.92°), and the offsets were both clockwise and counterclockwise. Since the representations were not centered on the training targets, each representation's amplitude did not coincide with the training target's measured adaptation index, but instead varied so as to best fit the data. Indeed, some amplitudes were larger (A = 1.39 for the 315° target) and others smaller (A = 0.69 for the 225° target) than unity, while mean of the amplitudes was 0.98±0.25, larger than the mean of the train target adaptation indices (0.92). That is, a best fit to the data found that the inferred representations were often much larger than the actual strength of the field.

As a further test of the standard model, we calculated the model's prediction for the uncorrected inter-gen curves. To do so, we used the standard model's premise that post-adaptation force measurements are a superposition of the estimated field and the baseline forces. Therefore, we added the baseline adaptation indices to the standard model's prediction of baseline corrected data for predictions of uncorrected data (Fig 4A). The standard model made precise predictions for the post-adaptation data (RMSE=0.109, AICc=-67.44). Next, we calculated asymmetries and compared them with their experimental counterparts in the inter-gen curves. The mean experimental asymmetry was significantly different from only one of the eight standard model predictions (target 0°: t-test, t(9)=3.46, p=0.0072). To be clear, the predicted asymmetry in this model was due to the combination of offset and amplitude, so the offset itself was not merely a fit to the experimental asymmetry.

There are a number of assumptions with the standard model that are cause for concern. For example, according to this model subjects should make straight reaches and apply negligible forces in baseline error-clamp trials, whereas we found systematic non-zero baseline perpendicular errors and error-clamp forces (Fig 2D). It is tempting to dismiss this critique, since baseline trajectories are of course not perfectly straight, so non-zero forces should be observed (owing to the limb's impedance). Yet this calls into question what error-clamp trials are actually measuring. If the intended movement is not perfectly straight, how does the limb's impedance influence error-clamp forces? And, since the limb's impedance changes with both posture, and commands, both of which change during testing and adaptation, how will these features influence the measured forces? Questions such as these must be addressed before we can properly infer what subjects learn.

Motivated by these questions, we examined a second model (see Limb Impedance Model, Methods), that did not assume the desired movements were planned to be straight, but in fact the movements observed during baseline. As a direct consequence of this assumption, the model must include limb stiffness and damping to predict channel forces. To allow for the fact that impedance changes with changes in commands, we included two sets of free parameters to scale the stiffness and damping before and after adapting to the field. To limit the number of model assumptions/parameters, the representation of the field was modeled just as before, except each representation was centered on actual the training direction (i.e. offset terms were absent). And as before, the model parameters were fit by maximizing the log likelihood of the generalization data.

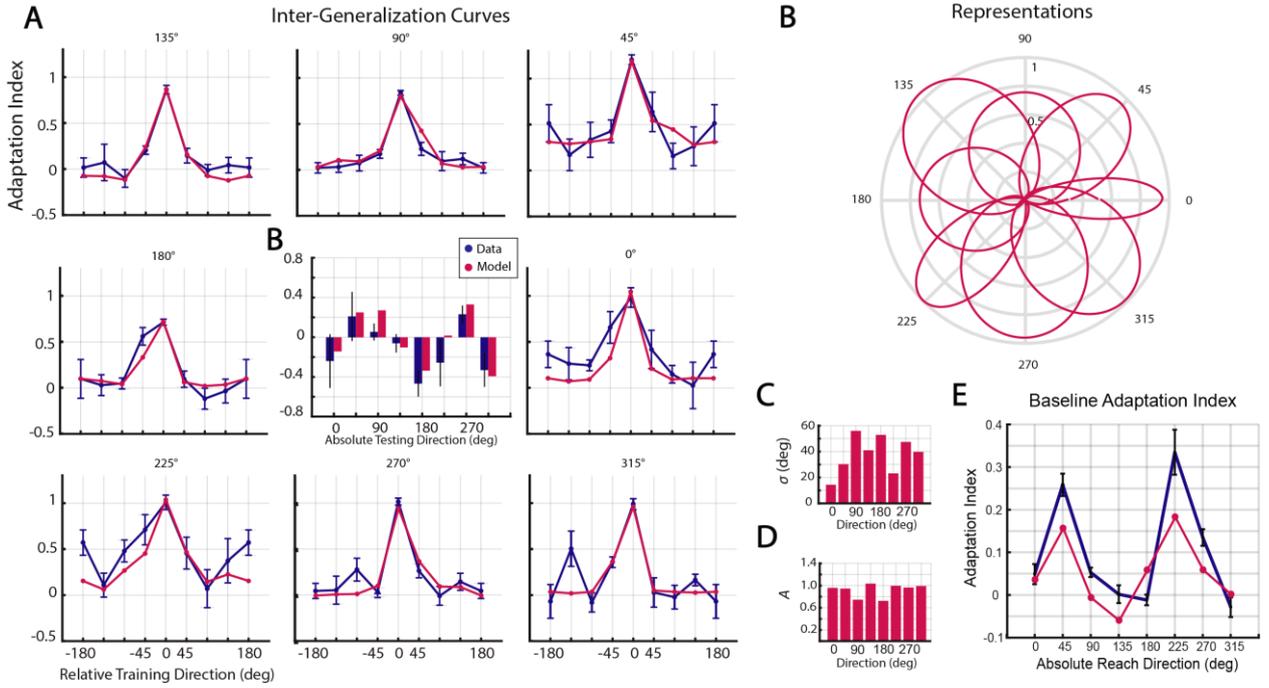

**Fig 5. Limb impedance model results.** (A) Experimental and model-predictions for post-adaptation inter-generalization data (blue and pink respectively). Predicted asymmetries together with their experimental counterparts (center panel). (B) The inferred representations are shown in polar coordinates for each of eight directions. The representation widths, σ (C) magnitudes, A (D). (E) Predicted baseline indices (pink) on top of the experimental baseline indices (blue) for eight directions.

In contrast with the standard model, this limb impedance model could predict baseline adaptation indices. We began by fitting the baseline measurements, assuming the amplitude of the representation at this time was zero, since there could be no estimate of the field. The model predicted forces when reaching through the perfectly straight error-clamps (owing to the limb's impedance), when the intended trajectory was slightly curved. The results accurately estimated the baseline measurements (Fig 5E), and the stiffness and damping of the limb, combined with the slight curvatures of baseline movements, resulted in non-zero error-clamp forces (baseline feedback gains: $\alpha k$= 0.7278, $\alpha b$= 0.0723, RMSE = 0.117, AICc = -25.64). Thus, the limb impedance model helped to explain why the baseline measurements indicated subjects were compensating for a non-existent clockwise curl field.

We then inferred the representations by fitting the post adaptation generalization data. This model yielded precise fits to the data (Fig 5A, feedback gains: $\alpha k$= 0.1406, $\alpha b$= 0.7108, RMSE=0.146, AICc=-42.26). As with the standard model, the inferred representations did differ across groups (Fig 5C, D). The width of the representations varied from 14.65° to 56.26° (average = 38.40°±14.56°), and the amplitudes varied from 0.75 to 1.04 (average = 0.92±0.12). However, relative to the previous model, these parameters were less variable when comparing the coefficient of variation (standard model: 39% for σ, and 24% for A; limb impedance model: 37% for σ and 13% for A).

The predicted asymmetries for the generalization curves were statistically indistinguishable from their experimental counterparts for all eight groups. Importantly, however, the representations were centered on the training targets. That is, while the underlying representation of the field is symmetric, the asymmetries in the generalization data are the natural result of how changes in reach direction and limb dynamics elicit variations in channel forces.

IV. DISCUSSION

Here we presented findings from a force field adaptation and generalization study exploring how systematic changes in limb configuration, and hence limb impedance, affect generalization. By examining adaptation and generalization across all eight standard directions our results were well-suited to examine the underlying representations. Behavioral results indicated the ability to generalize a learned force field was local relative to the practiced direction, for all eight directions. Furthermore, this ability was not symmetric for either the standard intra-generalization curves or the new inter-generalization curves that controlled for reach direction. These asymmetries were not consistent across directions and offered no easy explanation.

Using a standard interpretation for adaptation, a model-based analysis inferred the learned representations. By assuming the desired movements were minimum jerk, limb impedance has no effect on force channels and generalization measurements are a read-out of the estimated field. The model accurately fit the generalization curves and their asymmetries with Gaussian-like representations that shifted from side to side across training directions. A second model, including limb stiffness and damping and using subjects' baseline trajectories as their desired movements, also provided good fits to the data. It relied on fewer free parameters and used representations centered on the training directions.

The two models differed in their ability to explain the data, and their characterization of the learned representation of the force field. Despite the fact that all subjects adapted to the same field, the standard model found representations that varied across training directions. They were centered to the left or right of a training target, and had irregular widths and amplitudes that were at times larger than the actual field strength (and the measured adaptation index). The limb impedance model found more consistent representations with less variable parameters. These representations were centered on their training targets, and asymmetries in the generalization data were explained as due to the interactions of the force channel and limb impedance.

The standard model and its representations did provide a superior fit to the generalization data (RMSE=0.109, AICc=-67.44, versus RMSE=0.146, AICc=-42.26). The standard model, with its assumption of straight minimum jerk movements, cannot explain baseline behavior, whereas the impedance model accurately explained the pattern of baseline adaptation indices with fewer parameters (24 versus 18)..On the whole, based on these considerations and the more consistent parameters, we suggest the limb impedance model provides a more accurate description of the data and the underlying representations.

An issue with our limb impedance model is its potential sensitivity to the model parameters. Our nominal stiffness and damping terms were chosen from the literature [3, 20], but allowed for scaling (via $\alpha k$ and $\alpha b$). Based on our fit to the data, after adaptation the damping coefficient increased and the stiffness coefficient decreased, (baseline: $\alpha k$= 0.7278, $\alpha b$= 0.0723, post-adaptation: $\alpha k$= 0.1406, $\alpha b$= 0.7108). One might expect stiffness and damping to scale together with adaptation. Previous adaptation studies have found that the CNS is able to modify the magnitude, shape and orientation of the stiffness independent of the force needed to compensate for the perturbed dynamics [16]. We can only speculate that if these model parameters are accurately identifying changes in limb impedance, they suggest that stiffness and damping were modified to assist in making reaches in the field.

Other model parameters such as limb size, mass and inertia were chosen to represent an average subject. Based on our experience with the model small changes in these limb parameters do not qualitatively change the model's predictions. Furthermore, at various points of our investigation we have fit the models without all the subject's data and found qualitatively similar results. For these reasons we do not believe the results are sensitive to these parameters.

Limb impedance can influence generalization measurements in multiple ways. The impedance of the limb (e.g. inertia, damping, and stiffness) varies with changes in reach direction. To address this, we presented new inter-generalization curves that corrected for changes in reach direction during generalization. The impedance of the limb is also known to alter with changes in motor commands [16, 17]. Since adapting to a dynamical perturbation to the arm necessarily changes motor commands, our model allowed for changes in stiffness and damping by including scaling coefficients. Relative to the known complexities of musculoskeletal dynamics, these modifications were rudimentary and may not have captured salient features of how limb impedance influences generalization measurements. Future work can explore new experimental paradigms and models to more accurately infer what the brain learns when adapting.

Several of our findings are curious, if not altogether noteworthy, and merit future investigation. For example, despite all subjects adapting to the same field, the generalization curves were quite variable across training directions. We again note that reaches made in channel trials had no visual feedback to mitigate the effects of corrective, feedback driven changes in the movement. However, this may have contributed to the variability across reach directions and subjects. In future studies we intend to utilize more channel trials and use more subjects to address this concern.

Another finding that deserves re-examining were the positive adaptation indices found in baseline. Since all baseline reaches were curved slightly in the counterclockwise direction, this finding had a simple explanation in terms of the limb's non-zero impedance and the forces subjects should produce in the force channels. These measurements had a significant influence on our subsequent findings when generalization curves were corrected for baseline. If the baseline measurements had been negative, the overall strength, locality and asymmetries would have been different, yet presumably the underlying representation should have been identical. As above, more channel trials in the baseline blocks and further subjects would provide more data to scrutinize this finding.

Finally, chief among our follow-up questions is what, if any, is the influence of force field orientation (CW versus CCW) on our findings. Our original aim was the topic of generalization across directions, and as such we used the subjects we had to focus on that question. Since previous work on force field learning and generalization largely employed counterclockwise fields, we followed suit for easy and direct comparisons with the literature. Future work using a clockwise field would address a number of important issues. For example, if similar model fits were obtained, despite the fact that baseline-corrected measurements would effectively be weakened the generalization curves, this would demonstrate the robustness of these inferred representations. Additionally, we could ask if the variable representations we found remained constant, and were robust, or were in fact an artifact of the field's forces. Finally, additional results with a new field could validate the extent to which the limb's impedance could account for the generalization curves, independent of the field's orientation.

We end on a somewhat speculative note by pointing out a diverse line of work that has revealed similarities between motor learning and perceptual learning. There are examples from olfaction [26-32], tactile discrimination [33-36], and visual tasks [37-44] that find subjects' performance improves substantially with learning, but affords limited benefits in new situations. Although perceptual learning is very different from motor learning, there are some interesting conceptual similarities. Like motor learning, perceptual learning affords no conscious insight into how to improve performance on the task; subjects merely become better (in force field learning studies subjects often are unaware of what the field is doing, or even its existence). Perceptual learning can be described as using low-level sensory afferents to form a conscious percept, whereas motor learning can be described as a decision eliciting low-level motor efferents. In this sense it is tantalizing to speculate that motor and perceptual learning might be similar behaviors, albeit with information flowing in the opposite directions. If this were the case, then there is every reason to assume that what we learn about motor learning should be inherently informative for perceptual learning as well.

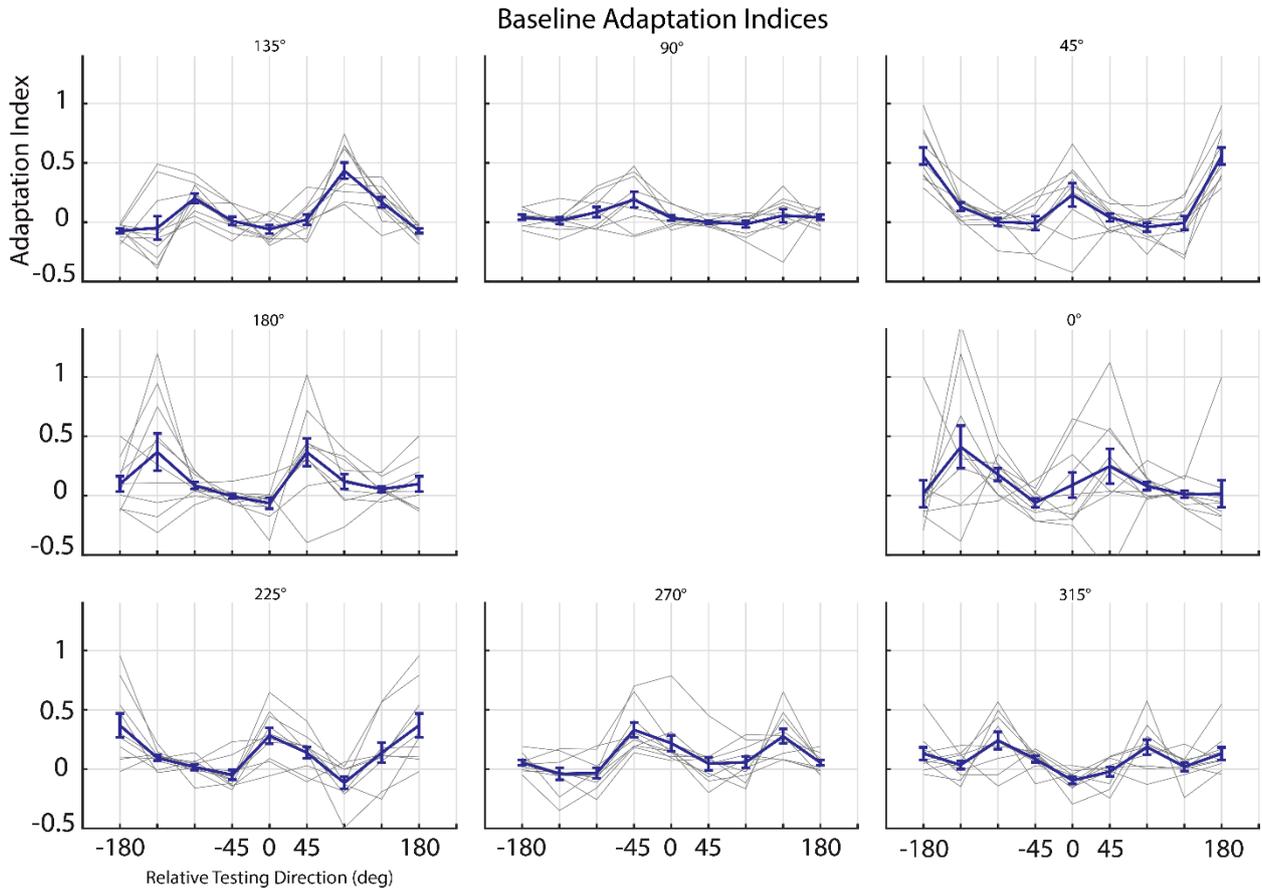

**Fig S1. Individual Baseline Adaptation Indices**. Average adaptation indices (gray) measured in baseline channel trials towards the eight target directions for individual subjects in each of eight groups (10 subjects per group). Across-subject's average for each group is also displayed (averages ± SEM, blue). Directions on top of each panel refers to the direction of the, yet to be presented, force field for each group later in adaptation trials.

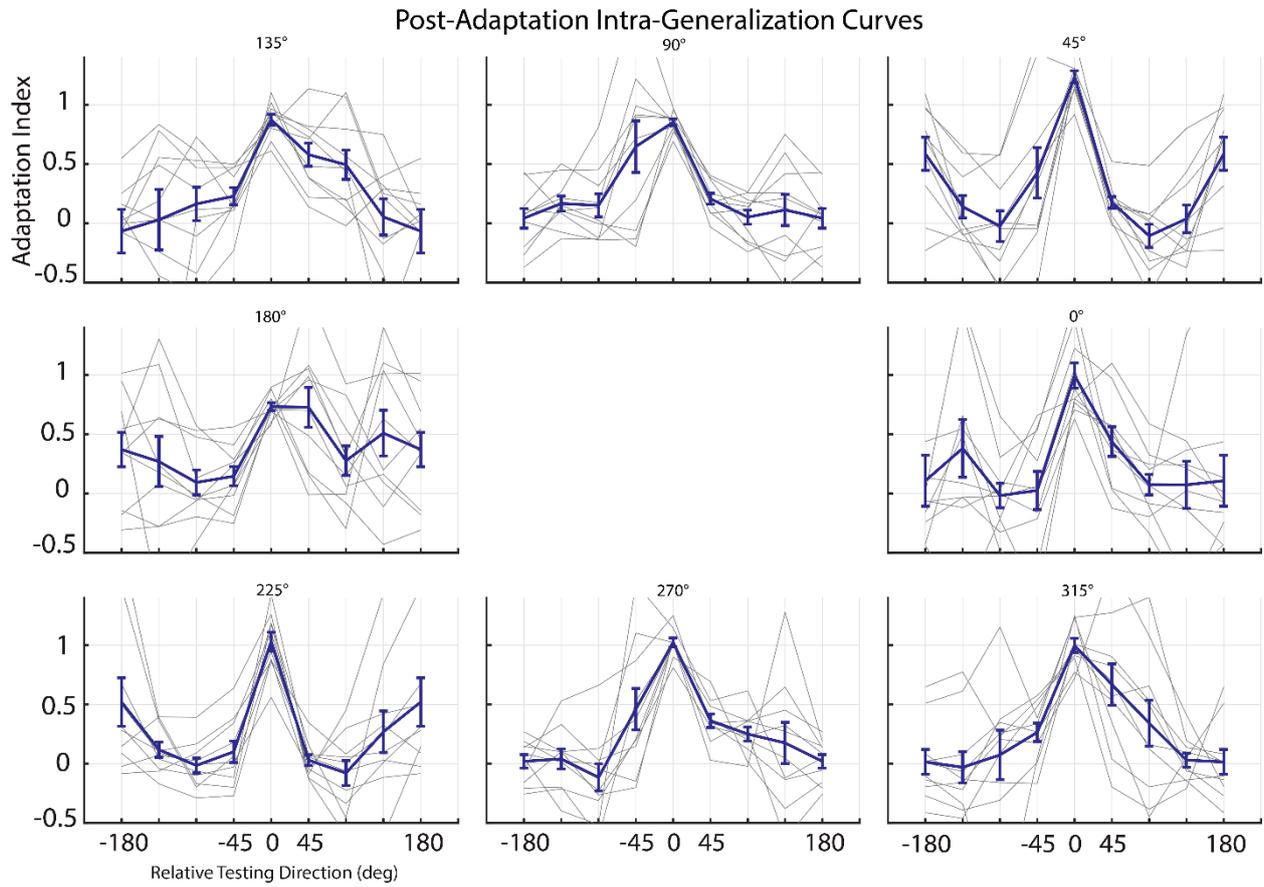

**Fig S2. Individual Post-Adaptation Intra-Generalization Curves**. Average adaptation indices (gray) measured in test block channel trials towards the eight target directions for individual subjects in each of eight groups (10 subjects per group). Across-subject's average for each group is also displayed (averages ± SEM, blue). Directions on top of each panel refers to the direction of the force field for each group in adaptation trials.

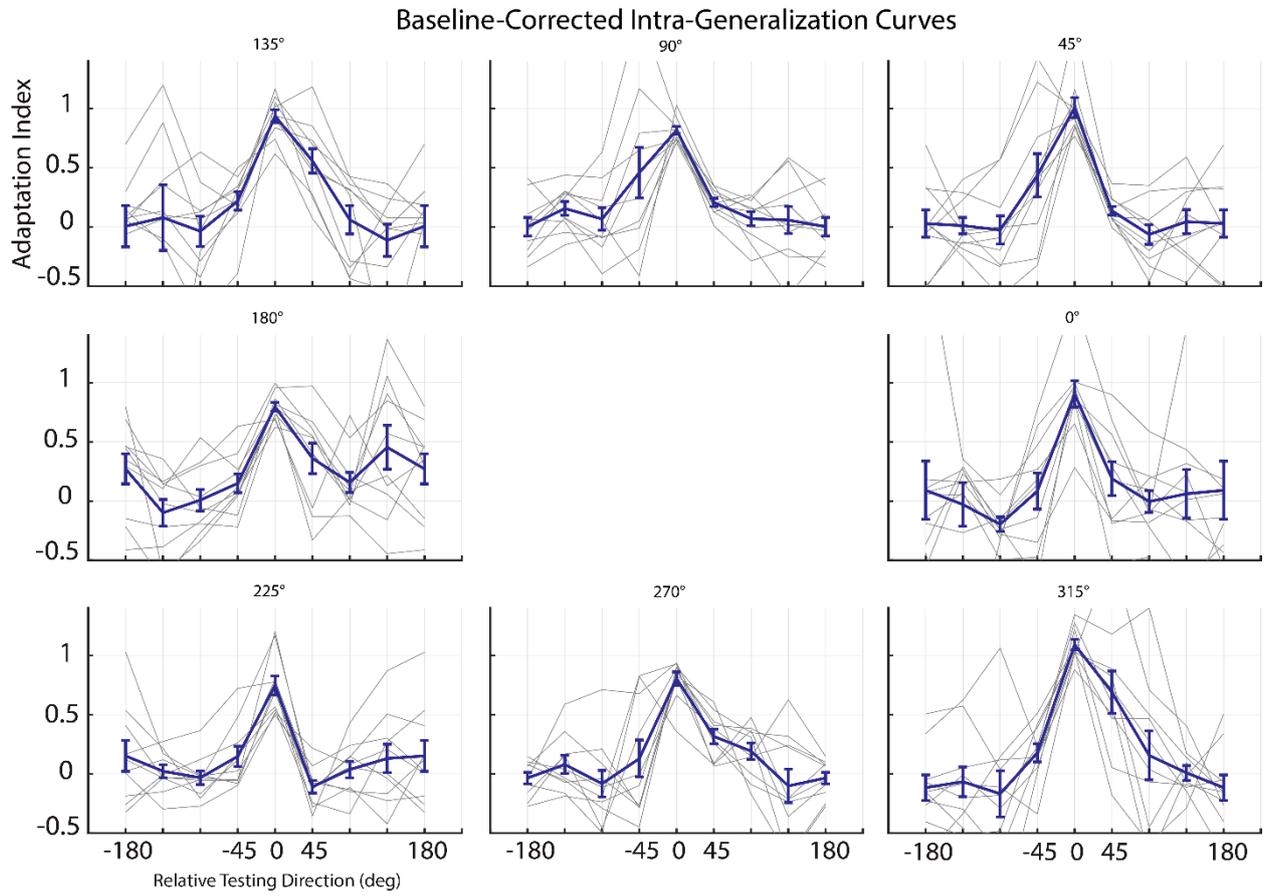

**Fig S3. Individual Baseline-Corrected Intra-Generalization Curves.** Average adaptation indices (gray) measured in test block channel trials towards the eight target directions for individual subjects in each of eight groups (10 subjects per group) corrected for the baseline indices (See Fig A-1). Across-subject's average for each group is also displayed (averages ± SEM, blue). Directions on top of each panel refers to the direction of the force field for each group in adaptation trials.